\documentclass[12pt]{article}
\usepackage{amsmath}
\usepackage{graphicx}
\usepackage{enumerate}
\usepackage{natbib}
\usepackage{url} 
\usepackage{amsfonts}
\usepackage{array}
\usepackage{booktabs}
\usepackage{float}
\usepackage{hyperref}
\usepackage{xcolor}

\definecolor{darkred}{RGB}{150,50,50}
\definecolor{brown}{RGB}{250,100,100}
\definecolor{green}{RGB}{000,150,100}
\definecolor{purple}{RGB}{250,000,180}

\usepackage{enumitem} 

\newcommand{\blind}{0}

\def\subPLM{_{\scriptscriptstyle \sf PLM}}
\def\subEHR{_{\scriptscriptstyle \sf EHR}}
\def\subINT{_{\scriptscriptstyle \sf INT}}
\begin{document}

\def\spacingset#1{\renewcommand{\baselinestretch}%
{#1}\small\normalsize} \spacingset{1}


\if1\blind
{
  \bigskip
  \bigskip
  \bigskip
  \begin{center}
    {\LARGE A Common {P}ipeline for Harmonizing {E}lectronic {H}ealth {R}ecord Data for {T}ranslational Research}\\

\vspace{5mm}

\end{center}
  \medskip
 \fi

\if0\blind
{
  \bigskip
  \bigskip
  \bigskip
  \begin{center}
    {\LARGE A Common {P}ipeline for Harmonizing {E}lectronic {H}ealth {R}ecord Data for {T}ranslational Research}\\

\vspace{5mm}

    Jessica Gronsbell, Vidul Ayakulangara Panickan, Doudou Zhou, Chris Lin, Thomas Charlon, Chuan Hong, Xin Xiong, Linshanshan Wang,  Jianhui Gao, Shirley Zhou, \\ Yuan Tian, Yaqi Shi, Ziming Gan, Tianxi Cai
\end{center}
  \medskip
} \fi

\bigskip
\begin{abstract}
\noindent
Despite the growing availability of Electronic Health Record (EHR) data, researchers often face substantial barriers in effectively using these data for translational research due to their complexity, heterogeneity, and lack of standardized tools and documentation. To address this critical gap, we introduce {\it{PEHRT}}, a common \underline{p}ipeline for harmonizing {EHR} data for \underline{t}ranslational research. PEHRT is a comprehensive, ready-to-use resource that includes open-source code, visualization tools, and detailed documentation to streamline the process of preparing EHR data for analysis. The pipeline provides tools to harmonize structured and unstructured EHR data to standardized ontologies to ensure consistency across diverse coding systems. In the presence of unmapped or heterogeneous local codes, PEHRT further leverages representation learning and pre-trained language models to generate robust embeddings that capture semantic relationships across sites to mitigate heterogeneity and enable integrative downstream analyses. PEHRT also supports cross-institutional co-training through shared representations, allowing participating sites to collaboratively refine embeddings and enhance generalizability without sharing individual-level data. The framework is data model-agnostic and can be seamlessly deployed across diverse healthcare systems to produce interoperable, research-ready datasets. By lowering the technical barriers to EHR-based research, PEHRT empowers investigators to transform raw clinical data into reproducible, analysis-ready resources for discovery and innovation.
\end{abstract}

\noindent%

\vfill

\newpage
\spacingset{1.9} 

\section{Introduction}
\label{sec:intro}
The growing availability of data from Electronic Health Records (EHRs) has transformed translational biomedical research. In the past decade, EHR data has been harnessed in a wide range of applications that have improved healthcare delivery and deepened our understanding of human health. These applications include dynamic risk prediction of diseases \citep{zhao2020incorporating, hong2023semi, hou2023risk, yang2023transformehr}, real-world treatment comparisons \citep{cheng2021robust, hou2023generate}, development of medical knowledge graphs \citep{hong2021clinical, li2024multi, tang2024leveraging}, and a broad range of genomic studies \citep{li2020electronic, xu2021quantitative, mccaw2024synthetic}. To fully leverage the potential of these applications, integrative analysis of multi-institutional EHR data is increasingly used to improve the generalizability of scientific findings, boost statistical power, and support the development of robust models for precision medicine \citep{cai2022consensus, guo2024multi, kundu2025privacy, li2025federated}. Multi-institutional research was particularly accelerated during the COVID-19 pandemic through the formation of international collaborative networks that conducted rapid, large-scale studies on COVID-19 treatment and management using federated analyses \citep{brat2020international, haendel2021national, vishwanatha2023community}. 

Progress notwithstanding, there are numerous barriers to effectively utilizing multi-institutional EHR data in translational applications. A key challenge is the lack of semantic interoperability across EHR systems, which results in substantial heterogeneity in  clinical documentation and medical coding practices \citep{hripcsak2013next, de2022semantic, sarwar2022secondary, yang2023machine, tang2024harnessing}. The foundation of any collaborative research study therefore rests on careful standardization of data elements across different data sources, a process known as {\it{data harmonization}}. Currently, there are no universally accepted or standardized procedures for harmonizing EHR data for an integrative analysis, despite the importance of such standards for ensuring the validity, transparency, and reproducibility of research findings \citep{ramakrishnaiah2023ehr}. The significance of proper data preparation became readily apparent during the COVID-19 pandemic when two high-profile multi-institutional studies published in {\it{The Lancet}} and {\it{The New England Journal of Medicine}} were retracted within months of publication \citep{mehra2020cardiovascular, mehra2020retraction}. In spite of passing some of the most rigorous peer reviews, the authors could not verify the data or processing procedures that underscored the validity of their conclusions. These incidents highlight the need for comprehensive and rigorous standards for harmonization to ensure the scientific integrity and credibility of collaborative research. 

To address the barriers that researchers face in using complex and heterogeneous EHR data for biomedical research, we developed {\it{PEHRT}}, a common \underline{p}ipeline for harmonizing \underline{EHR} data for \underline{t}ranslational research. Rather than introducing novel algorithms, PEHRT brings together a comprehensive collection of existing tools, methods, and software components, along with visualization utilities and detailed documentation to help researchers efficiently transform raw EHR data into analysis-ready formats. Motivated by our experience in federated EHR networks such as the Consortium for Clinical Characterization of COVID-19 by EHR (4CE) and the AIM-AHEAD program \citep{brat2020international,vishwanatha2023community}, PEHRT includes modules for mapping structured and unstructured EHR data to standardized coding systems, generating representations to mitigate heterogeneity across sites, and curating research-ready multi-institutional datasets without sharing individual-level data. The resulting resources can be used for diverse applications, including knowledge graph construction, phenotyping, predictive modeling, clinical studies, and federated learning and are accessible through fully documented open-source \texttt{R} and \texttt{Python} packages, web-based APIs for data visualization, and a user-friendly tutorial (\url{https://celehs.github.io/PEHRT/}
).

The remainder of this paper is organized as follows. Section \ref{sec:comparison} provides a review of related literature. Section \ref{sec:overview} details the two PEHRT modules: (1) data pre-processing and (2) representation learning. Section \ref{sec:conc} discusses the implications and limitations of our work. A real-world example illustrating the utility and execution of PEHRT in several downstream tasks using EHR data from multiple healthcare systems is provided in the Supplementary Materials.

\section{Existing Tools and Gaps}
\label{sec:comparison}
Existing research has primarily focused on specific aspects of EHR data preparation within individual institutions, including data cleaning, data standardization, medical code aggregation, and quality assessment \citep{pathak2013normalization, makadia2014transforming, fhir_v5_index, OHDSIDataStandardization}. Data cleaning involves transforming and normalizing raw EHR data, conducting exploratory data analysis, detecting anomalies, and scaling and transforming data \citep{hong2019developing, mandyam2021cop, ramakrishnaiah2023ehr, muse2024protocol}. Standardization involves mapping raw data to common data models and aligning medical codes with established medical coding systems or ontologies. Open-source tools, including Electronic Health Record Quality Control (EHR-QC) \citep{ramakrishnaiah2023ehr}, the Cohort Migrator Toolkit (CMToolkit) \citep{CMToolkit}, and the Observational Health Data Sciences and Informatics (OHDSI) network’s Themis, are available to convert data to the widely used Observational Medical Outcomes Partnership-Common Data Model (OMOP-CDM) \footnote{\url{https://www.ohdsi.org/data-standardization/}}. 

Following standardization, medical codes are aggregated or ``rolled-up'' into broader medical concepts to represent clinically meaningful variables as disaggregated data are often too granular for research purposes. For example, codes within standard medical coding systems, such as the International Classification of Diseases (ICD) for diagnoses, are typically rolled-up into higher-level concepts using established ontologies. Code roll-up can be done manually or with machine learning approaches \citep{zhang2019automated}. Lastly, quality assessment of EHR data is conducted using established criteria or open-source tools, such as the Automated Characterization of Health Information at Large-scale Longitudinal Evidence Systems (ACHILLES) or the Data Quality Dashboard (DQD) from the OHDSI network \citep{lewis2023electronic}. ehrapy, a modular, open-source Python framework, is the only end-to-end tool currently available for the curation and analysis of EHR data \citep{heumos2024open}. The framework is designed for exploratory data analysis and consists of modules for data pre-processing and ontology mapping as well as analysis tools for causal inference, survival analysis, and patient stratification. 

In spite of the large volume of work devoted to EHR data preparation, significant gaps remain when working with multi-institutional EHR data \citep{aminoleslami2024ehrs}. Existing tools designed for data from a single institution fail to address the variability in coding practices across institutions. For example, many health systems use local medical codes (i.e., codes specific to their system) that are not mapped to standardized coding systems. Traditionally, standardization has been achieved by mapping local codes to standard coding systems, either manually or using automated tools like Medication Extraction and Normalization (MedXN) \citep{sohn2014medxn, ramakrishnaiah2023ehr}. Recently, advances in Large Language Models (LLMs) and representation learning have facilitated the generation of semantic embeddings, which are vector representations that capture the meanings of EHR codes and their relationships. Embeddings can be used to significantly enhance the efficiency and accuracy of data standardization, which is a critical aspect of preparing multi-institutional data. However, available pipelines lack user-friendly modules for local code standardization and do not leverage representation learning for harmonization.  

\textit{PEHRT} was developed to address this gap by providing a comprehensive, user-oriented resource. It integrates scalable, adaptable methods, software, visualization tools, and detailed protocols into a unified pipeline that enables researchers to efficiently transform raw EHR data into interoperable, analysis-ready datasets. Specifically, PEHRT contains tools to harmonize structured and unstructured EHR data to standardized ontologies, along with resources that use representation learning and LLM-based embeddings to mitigate heterogeneity and align local, unmapped codes across institutions. The pipeline also supports cross-institutional co-training through representation sharing to facilitate collaborative research without  individual-level data exchange. Additionally, PEHRT provides detailed, practical guidance for code roll-up, pre-processing, and integration of diverse structured and unstructured data types (e.g., diagnoses, medications, laboratory results, procedures, and clinical notes). Together, these resources lower the technical barrier to EHR-based research, allowing investigators to focus on scientific discovery rather than data harmonization.

\section{Overview of PEHRT}
\label{sec:overview}
PEHRT consists of two modules: (1) data pre-processing and (2) scalable representation learning for data harmonization. The input to PEHRT is a set of original EHR datasets from multiple institutions and the output is a harmonized, research-ready dataset \footnote{While developed for integrative analysis, PEHRT can also be used to prepare datasets within a single institution by skipping Step 2.3 (Joint multi-institution EHR embedding training).}. One of our key contributions is an online tutorial (see Figure~\ref{fig:overview}), which guides users through each step of PEHRT using publicly available EHR data from the Medical Information Mart for Intensive Care IV (MIMIC-IV) database \citep{johnson2023mimic}. 

Prior to utilizing PEHRT, it is necessary for researchers to familiarize themselves with their EHR data sources, including relevant documentation, data schema, and coding systems. PEHRT is designed to accommodate EHR data stored in relational databases and the original EHR datasets need not be represented in a common data model (e.g., OMOP). Additional details about the equipment and software requirements can be found in the tutorial introduction (\url{https://celehs.github.io/PEHRT/}).  

\begin{figure}[H]
    \centering
    \fbox{\includegraphics[scale = 0.5]{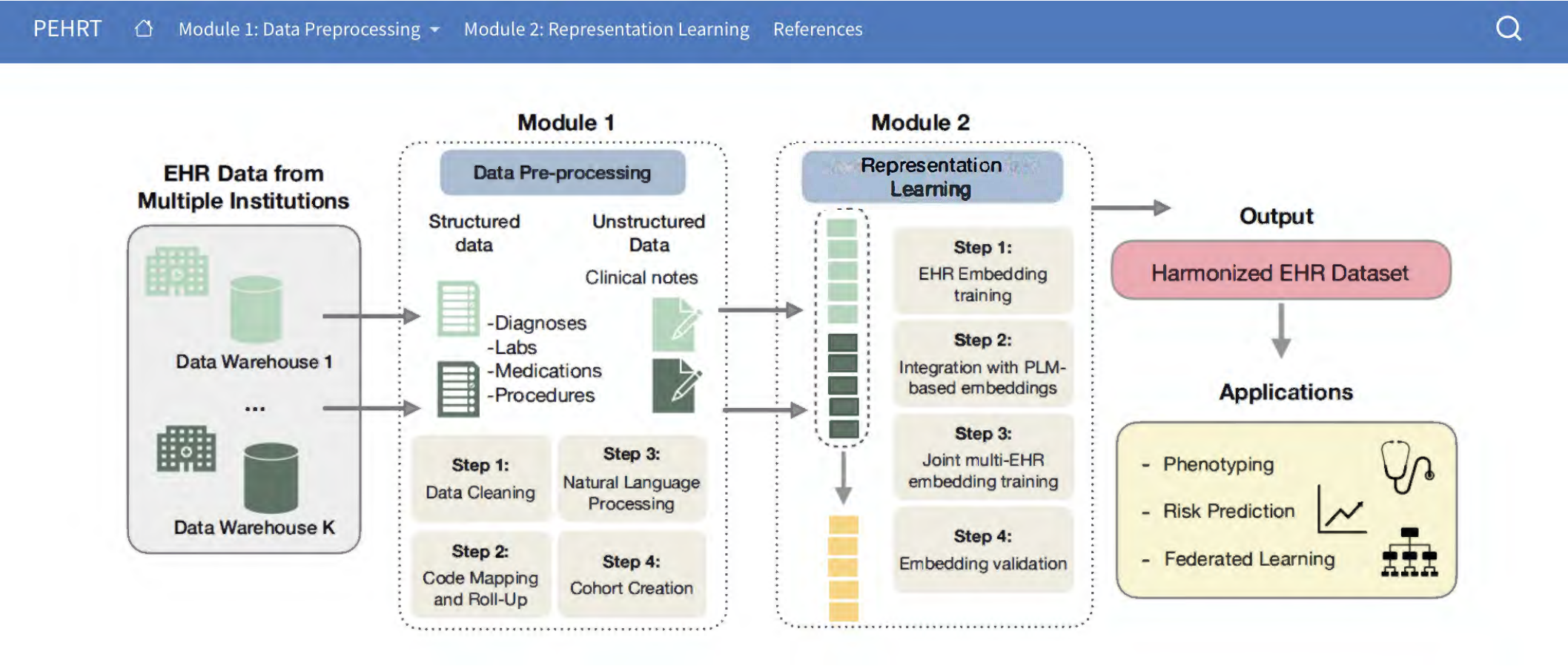}} 
    \caption{PEHRT enables users to prepare a harmonized, research-ready dataset with two modules for data pre-processing and representation learning. Each step of PEHRT is detailed in our user-friendly tutorial and supported by open-source software and web APIs for data visualization.}
    \label{fig:overview}
\end{figure}

\subsection{Module 1: Data Pre-processing}
\label{sec:overview_module1}%
Data pre-processing is a meticulous process that involves several sub-steps, including (1.1) data cleaning, (1.2) code mapping and roll-up, (1.3) Natural Language Processing (NLP) of free-text data, and (1.4) cohort creation. Pre-processing is performed on each EHR dataset that is input to PEHRT with the goals of transforming the data to a more usable format and standardizing data across institutions. PEHRT enables processing of a broad range of structured data, including diagnostic codes, medication prescriptions, laboratory tests, and procedure codes, as well as unstructured data in the form of clinical notes.

\href{https://celehs.github.io/PEHRT/EHR-Pipeline-Part2.html#step-3-cleaning-the-data}{\textbf{Step 1.1: Data Cleaning.}}
PEHRT employs a multi-step data cleaning process to enhance the quality of noisy and fragmented EHR data. Following the merging of relevant data tables, variables irrelevant to downstream analytical tasks are excluded to improve computational efficiency and reduce memory usage. Next, duplicate records, which are common in EHR data, are identified and removed. These may include exact duplicates or semantic duplicates, which are distinct records that have equivalent information (e.g., records with different medical codes corresponding to the same clinical concept). De-duplication is also performed after code mapping and roll-up (as described in Step 1.2) to ensure that all redundant information is eliminated prior to analysis. Additionally, erroneous data entries, which are frequent in EHR datasets, are carefully assessed. Obvious errors such as impossible or out-of-range dates are filtered out. However, other data elements, such as laboratory measurements, may appear erroneous due to differing units. In such cases, domain knowledge is applied to normalize units and prevent misclassifying valid data. Lastly, records missing critical elements, such as dates or event information, are removed. For large-scale EHR datasets, we recommend batch processing, which is demonstrated in our online tutorial.

\href{https://celehs.github.io/PEHRT/EHR-Pipeline-Part3.html}{\textbf{Step 1.2: Code Mapping and Roll-Up.}} 
Medical codes are often too specific for research studies. To address this issue, PEHRT standardizes codes by mapping them to recognized coding systems and then aggregating or ``rolling-up'' the codes into higher-level categories across the domains of interest. Code roll-up provides consistency across diverse EHR datasets while also ensuring data is at an appropriate level of granularity for analysis. PEHRT contains detailed instructions for code roll-up across four domains using established ontologies: diagnoses, medications, laboratory tests, and procedures.

For diagnoses, we use the Phenotype Code (PheCode) hierarchy for ICD codes \citep{denny2010phewas}. The PheCode hierarchy provides 1,875 integer, 1-digit, and 2-digit level codes that capture a wide range of disease conditions while avoiding excessive granularity. The hierarchy also provides parent-child relationships that characterize associations between PheCodes. While there are several versions, we use Phecode v1.2 to map both ICD-9-CM (ICD, Ninth Revision, Clinical Modification) and ICD-10-CM codes for illustration in the tutorial. CM codes are used across the U.S. for billing purposes. 

For medications, we recommend rolling up Prescription Normalized Names and Codes (RxNorm) codes to ingredient-level codes unless the study specifically requires dosage information. For studies involving drug classes, ingredient-level codes can be further rolled up according to existing ontologies including the Anatomical Therapeutic Chemical (ATC) classification \footnote{\url{https://www.who.int/tools/atc-ddd-toolkit/atc-classification}}, the  Accrual to Clinical Trials (ACT) ontology, or the Veterans Affairs (VA) drug class \footnote{\url{https://www.nlm.nih.gov/research/umls/rxnorm/sourcereleasedocs/vandf.html}}, depending on the requirements of the study.  Additionally, some medications may be coded as procedures due to the method of administration or billing requirements, and it is necessary to map these codes to relevant RxNorm codes. For instance, J codes from the Healthcare Common Procedure Coding System (HCPCS) are often used for injectable drugs. J codes can be mapped to RxNorm by leveraging crosswalks between coding systems with the Unified Medical Language System (UMLS).

For laboratory tests, we use the Logical Observation Identifiers Names and Codes (LOINC) hierarchy. Measurements for the same analyte can vary due to differences in the specimen, time of measurement, method, or scale and result in multiple LOINC codes. We therefore recommend rolling up to the lowest level of LOINC part \citep{mcdonald2003loinc}. PEHRT includes a searchable and downloadable web API to assist users in visualizing the LOINC hierarchy, as well as ICD and RxNorm \footnote{\url{https://shiny.parse-health.org/hierarchies/}}. However, it is important to note that PEHRT only supports the usage of laboratory codes. Preparing laboratory result data is an involved process that requires informatics experts familiar with the EHR datasets of interest as it requires unit harmonization and specialized quality control.  

Unfortunately, few established hierarchies exist for procedure codes. PEHRT illustrates how to roll up procedure codes into categories according to the Clinical Classification Software (CCS) \footnote{\url{https://hcup-us.ahrq.gov/toolssoftware/ccs_svcsproc/ccssvcproc.jsp}}. As many institutions use both Current Procedural Terminology (CPT) and ICD codes, it is critical to include both when rolling-up procedure codes. 

\href{https://celehs.github.io/PEHRT/EHR-Pipeline-Part4.html}{\textbf{Step 1.3: Natural Language Processing.}} When free-text clinical notes are also available, one may employ NLP tools to extract clinical concepts from the notes by identifying and mapping terms such as diseases, symptoms, and medications to Concept Unique Identifiers (CUIs) in the UMLS. Existing NLP software tools like NILE \citep{yu2013nile}, cTAKES \citep{savova2010mayo}, or MetaMap \citep{aronson2001effective} enable this extraction, allowing for semantic analysis and structured representation of clinical text, which is then integrated into the dataset for downstream analysis. We previously introduced a pipeline for EHR phenotyping \citep{zhang2019high}, which contains detailed steps for running NLP as well as an online tutorial (\url{https://celehs.github.io/PheCAP/}).

\href{https://celehs.github.io/PEHRT/EHR-Pipeline-Part5.html}{\textbf{Step 1.4: Cohort Creation.}}
EHR-based studies are typically conducted on a group of patients who meet specific inclusion/exclusion criteria, such as those with certain diagnoses, medications, or procedures. PEHRT streamlines cohort identification by leveraging the standardized and rolled-up codes from Step 1.2. For example, when the cohort is identified based on a particular disease diagnosis, a common strategy for identifying the patient cohort is to use relevant ICD codes \citep{shivade2014review, banda2018advances, yang2023machine}. However, ICD codes can be overly granular, which often leads to different studies using inconsistent sets of ICD codes to capture the condition of interest. To address this issue, PEHRT utilizes PheCodes from Step 1.2 to identify patients associated with the condition of interest. Following cohort identification, PEHRT then aggregates the rolled-up codes from Step 1.2 and the CUIs from Step 1.3. For studies involving temporal analysis, we recommend further aggregating patient-level longitudinal data into time windows, such as monthly counts or averages. For example, for chronic conditions like rheumatoid arthritis, monthly aggregation typically provides sufficient granularity while simplifying downstream analysis. 

\subsection{Module 2: Representation Learning}
\label{sec:overview_module2}
Following data pre-processing, representation learning is used to develop institution-specific embeddings and a joint embedding that leverages the data across institutions without the sharing of individual-level data. PEHRT also incorporates embeddings from pre-trained language models (PLMs) to further enhance the quality of the learned data representations. The embeddings can be used for a variety of downstream tasks within and across institutions, including mapping local codes to recognized coding systems, knowledge graph construction, phenotyping, and predictive modeling (see Supplementary Materials for examples) \citep{hong2019developing, xiong2023knowledge, zhou2021multi}. The module consists of four sub-steps: (2.1) EHR embedding training, (2.2) PLM-based embedding generation, (2.3) joint embedding training, and (2.4) embedding validation. 

\href{https://celehs.github.io/PEHRT/m2.html#step-2.1-embedding-training}{\textbf{Step 2.1: EHR Embedding Training.}}  PEHRT first generates institution-specific EHR embeddings from summary-level data using the Singular Value Decomposition of the Pointwise Mutual Information (PMI-SVD) algorithm \citep{beam2020clinical}. This method factorizes the PMI matrix constructed from co-occurrence counts of the codes and CUIs from Module 1. As a variant of the widely adopted word2vec algorithm \citep{levy2014neural}, PMI-SVD has proven to be highly effective in learning meaningful and interpretable clinical embeddings.

The algorithm consists of three steps. First, a co-occurrence matrix ${\bf{C}} = [C(w, c)]$ is constructed, where each element $C(w, c)$ represents the number of patients in which a target code or CUI $w$ co-occurs with a context code or CUI $c$ within a predefined time window (e.g., $30$ days). This matrix captures the local context of clinical concepts and provides a foundation for computing semantic similarity. We previously developed an algorithm for efficient co-occurrence computation as calculating ${\bf{C}}$ at scale is computationally intensive \citep{hong2021clinical, rusheniii_LargeScaleClinicalEmbedding, gan2025arch}. This algorithm is integrated into PEHRT software to enable scalable training of EHR embeddings for large datasets. 

Next, the co-occurrence matrix is used to calculate the shifted positive PMI (SPPMI) matrix, which represents the relationships among codes and CUIs. The SPPMI matrix is defined as
$$\text{SPPMI}(w, c) =  \max \left\{ \log \frac{C(w, c) \cdot |D|}{C(w, \cdot) C(\cdot, c)}  -\log k, 0 \right\}$$
where $C(w, \cdot)$ is the row sum of $C(w,c)$, $C( \cdot, c)$ is the column sum of $C(w,c)$, $|D|$ is the total sum of the co-occurrence, and $k$ is the negative sampling rate. We have found that the embedding quality is generally not sensitive to the length of the time window, but is best when $k = 1$ \citep{hong2019developing}. Lastly, the SPPMI matrix is decomposed with its rank-$d$ SVD, represented as ${\bf{Q}}_{d} {\bf{\Lambda}} {\bf{Q}}_{d}$. PEHRT outputs the $d$-dimensional embedding vectors as $ \mathbf{X}\subEHR = {\bf{Q}}_d {\bf{\Lambda}}^{1/2}$. To select $d$, we recommend retaining a large amount of variation in the SVD (e.g., 95\%) by evaluating the eigenvalue decay \citep{hong2021clinical, hou2023generate}. Alternatively, $d$ can be selected by maximizing the area under the receiver operating characteristic curve (AUC) for discriminating between pairs of codes and CUIs with known relationships against randomly selected pairs (see Step 2.3 for further details).

\href{https://celehs.github.io/PEHRT/m2.html#step-2.2-plm-based-embeddings}{\textbf{Step 2.2: PLM-based Embeddings.}} PEHRT generates a second set of embeddings using PLMs to leverage the textual descriptions of codes and CUIs (e.g., CUI C0020538: blood pressure, high). PLMs are trained on large text corpora and, in some cases, further fine-tuned with biomedical knowledge sources such as PubMed articles, clinical notes, and knowledge graphs. The PLM-based embeddings complement the EHR embeddings from Step 2.1 that capture the meaning of codes and CUIs based on their usage within a healthcare institution. PEHRT includes instructions for generating embeddings from several commonly used PLMs, including the \textbf{C}ross-lingual kn\textbf{o}wledge-infused me\textbf{d}ical t\textbf{er}m embedding (CODER) \citep{yuan2022coder}, \textbf{S}elf-\textbf{A}ligned \textbf{P}re-trained \textbf{B}idirectional \textbf{E}ncoder \textbf{R}epresentations from \textbf{T}ransformers (SapBERT) \citep{liu2021self}, PubMedBERT \citep{pubmedbert}, and BERT for Biomedical Text Mining (BioBERT) \citep{lee2020biobert}.

When working with data from a single institution, PLM-based embeddings can be combined with the PMI-SVD embeddings from Step 2.1 to improve overall quality. A simple yet effective approach is to form a weighted concatenation of the two embeddings, with the weight adjusted to the specific downstream task. Specifically, we define 
\begin{equation*}
    \mathbf{X}\subINT = [ w \mathbf{X}\subEHR, (1-w) \mathbf{X}\subPLM ]  
\label{eq:w}
\end{equation*}
where $\mathbf{X}\subPLM$ is the PLM-based embedding, $w \in [0,1]$, and both $\mathbf{X}\subEHR$ and $\mathbf{X}\subPLM$ are normalized so that each row has unit norm. The final integrated embedding is then obtained by normalizing each row of $\mathbf{X}\subINT$ to unit norm. This integration leverages the complementary strengths of the two EHR and PLM-based embeddings: PMI-SVD effectively identifies clinically related codes and CUIs (e.g., drug–disease pairs) while PLMs capture semantic similarities from textual descriptions \citep{liu2021self, zhou2022multiview}.

\href{https://celehs.github.io/PEHRT/m2.html#step-2.3-joint-multi-institution-ehr-embedding-training}{\textbf{Step 2.3: Joint Embedding Training.}} To leverage data across institutions, PEHRT uses the BONMI algorithm \citep{zhou2021multi} to derive a shared representation of EHR codes and CUIs by aligning and completing institution-specific SPPMI matrices. Briefly, BONMI constructs an aggregated matrix covering all unique codes and CUIs, assigning weighted averages to overlapping pairs and marking others as missing. The weights are based on data quality using user-defined or data-driven metrics and the missing values are imputed by aligning institution-specific embeddings via orthogonal transformations. The completed matrix is then factorized with an SVD to generate the joint embedding, with rank selection as in Step 2.1. The joint embedding can be integrated with PLM-based embeddings through weighted concatenation as described in Step 2.2. 

\href{https://celehs.github.io/PEHRT/m2.html#step-2.4-embedding-validation}{\textbf{Step 2.4: Embedding Validation.}} To evaluate the quality of the trained embeddings, PEHRT provides simple metrics quantifying their performance in discriminating between concept pairs with known relationships against randomly selected pairs. The relationships can be curated from existing ontologies and UMLS. For each pair under consideration, the cosine similarity of the corresponding embedding vectors is calculated to measure their degree of relatedness. Embedding quality is then quantified based on the AUC of the cosine similarity in distinguishing between the related and random pairs (i.e., the probability that a randomly selected related pair will have a higher cosine similarity than a randomly selected random pair). These metrics can be used to evaluate the performance of the institution-specific EHR embeddings, the PLM-based embeddings, as well as the joint embedding. We have found that the joint embeddings generally achieve the highest performance in a wide variety of applications, but recommend comparing their performance with PLM-based embeddings and EHR embeddings for thorough evaluation \citep{xiong2023knowledge, gan2025arch, zhou2025representation}.

\section{Conclusion}\label{sec:conc}
Data harmonization is essential for ensuring the validity, transparency, and reproducibility of multi-institutional EHR-based research. However, significant heterogeneity across data sources complicates harmonization and no comprehensive and standardized procedures currently exist to address this challenge. To fill this gap, we introduced PEHRT, a common pipeline for harmonization of EHR data for translational applications. PEHRT operates entirely on summary-level data to preserve data privacy and is designed for easy implementation through our online tutorial and suite of resources. Our pipeline also leverages representation learning and PLMs to streamline data standardization and support a wide range of applications to advance clinical research and practice, including phenotyping, knowledge graph construction, and federated learning \citep{hong2021clinical, xiong2023knowledge, gan2025arch, zhou2025representation}. An example application of PEHRT in several integrative predictive modeling tasks is detailed in the Supplementary Materials to further support users in applying the pipeline for their own purposes.

However, our work is not without limitations. First, the EHR embeddings rely on longitudinal data to compute the co-occurrence matrix. Including additional data sources without repeated observations (e.g., genomic data) requires different methodological approaches. Second, incorporating temporal information (e.g., one diagnosis precedes another) into the EHR embeddings has the potential to improve the applicability of PEHRT. For example, such embeddings would be particularly relevant to studies aimed at identifying comorbidities and risk factors for a disease. Third, the joint embedding training can potentially be improved by incorporating the relative quality of each data source. Lastly, a future direction is to further optimize PEHRT to allow for incremental updating over time using zero-shot or few-shot learning.

\section*{Data Availability Statement}

The MIMIC-IV dataset analyzed in this study is available to credentialed researchers who complete the required data use agreement, which can be accessed at \url{https://physionet.org/content/mimiciv/3.1/}. The additional electronic health record datasets used in the supplemental example are not publicly available due to patient privacy and institutional data-use restrictions. All analysis code necessary to reproduce the MIMIC-IV results is openly available in our online tutorial at \url{https://celehs.github.io/PEHRT/}.

\newpage

\if1\blind
{
  \bigskip
  \bigskip
  \bigskip
  \begin{center}
    {\LARGE Supplementary Materials for \\
  \LARGE A Common {P}ipeline for Harmonizing {E}lectronic {H}ealth {R}ecord Data for {T}ranslational Research}
\end{center}
  \medskip
} \fi

\if0\blind
{
  \bigskip
  \bigskip
  \bigskip
  \begin{center}
    {\LARGE Supplementary Materials for \\A Common {P}ipeline for Harmonizing {E}lectronic {H}ealth {R}ecord Data for {T}ranslational Research}\\

\vspace{5mm}

    Jessica Gronsbell, Vidul Ayakulangara Panickan, Doudou Zhou, Chris Lin, Thomas Charlon, Chuan Hong, Xin Xiong, Linshanshan Wang,  Jianhui Gao, Shirley Zhou, \\ Yuan Tian, Yaqi Shi, Ziming Gan, Tianxi Cai
\end{center}
  \medskip
} \fi

\def\spacingset#1{\renewcommand{\baselinestretch}%
{#1}\small\normalsize} \spacingset{1}

\section{Example: Using PEHRT for Predictive Modeling}\label{Section:Pred-Modeling}

\subsection{Overview \& Data Sources}
Here we illustrate the two modules of PEHRT as well as its application in downstream integrative predictive modeling tasks. Specifically, we detail the data pre-processing for the MIMIC-IV dataset \citep{johnson2023mimic} that is utilized in the online tutorial (\url{https://celehs.github.io/PEHRT/}) and the training of joint embeddings using the MIMIC-IV data as well as data from the Mass General Brigham (MGB), the Veterans Health Administration (VA), Boston Children's Hospital (BCH), and the University of Pittsburgh Medical Center (UPMC). In our analysis, the MGB EHR data contains 2.5 million patients from 1998 to 2018. The VA Corporate Data Warehouse (CDW) aggregates data from $150$ VA facilities into a single data warehouse, with records from 1999 to 2019 covering $12.6$ million patients. The BCH data contains $251$K patients from 2009 to 2022 and the UPMC EHR data includes $95$K patients from 2004 to 2022, focusing on individuals with at least one occurrence of ICD codes related to Alzheimer’s disease and dementia or multiple sclerosis. The MIMIC-IV dataset contains $65$K ICU admissions and over $200$K emergency department admissions at Beth Israel Deaconess Medical Center in Boston, Massachusetts, spanning 2008 to 2019. 

\subsection{Module 1: Data Pre-processing}
We used Module 1 of PEHRT to pre-process structured EHR data from all participating institutions. All of the pre-processing steps are demonstrated in our online tutorial using the MIMIC-IV dataset. The tutorial provides guidance on setting up the workspace, obtaining access to MIMIC-IV, and getting familiar with the dataset (\href{https://celehs.github.io/PEHRT/EHR-Pipeline-Part1.html}{link}). 

The inputs to Module 1 are the original data tables from the MIMIC-IV database and the output is a pre-processed dataset. The pre-processing begins with data cleaning, which involves assessing missingness, standardizing schema across tables, removing irrelevant and redundant information, and constraining the data to the relevant time window (\href{https://celehs.github.io/PEHRT/EHR-Pipeline-Part2.html}{link}). 

Next, we provide step-by-step instructions to roll-up diagnosis codes, procedure codes, and medications (\href{https://celehs.github.io/PEHRT/EHR-Pipeline-Part3.html}{link}). For EHR text processing, we generally recommend NILE software, but utilize a lightweight custom NLP library for illustrative purposes in the tutorial (\href{https://celehs.github.io/PEHRT/EHR-Pipeline-Part4.html}{link}). We also demonstrate how to refine the data to a cohort for a specific analysis, using asthma as an example  (\href{https://celehs.github.io/PEHRT/EHR-Pipeline-Part5.html}{link}).  

\subsection{Module 2: Representation Learning}
\subsubsection{Training EHR, PLM-based, and Joint Embeddings}
Within each institution, we obtained EHR embeddings from the PMI-SVD algorithm detailed in Module 2 using PheCodes, CCS categories, RxNorm codes, LOINC codes, and institution-specific local codes. Table~\ref{tab:code_summary} summarizes the number of unique codes across the $5$ institutions and the various coding domains. As expected, substantial heterogeneity exists in terms of the number of unique codes within each domain. We also obtained PLM-based embeddings using CODER, SapBERT, BioBERT, PubMedBERT, BGE, OpenAI's text-embedding-3-small model \footnote{BGE and OpenAI embeddings can be included in the \texttt{config\_dict} in Module 2. For OpenAI, the user must use their own API key.}, and the joint EHR embedding obtained using the BONMI algorithm described in Module 2. An illustrative example using the MIMIC-IV data is provided in our online tutorial (\href{https://celehs.github.io/PEHRT/m2.html}{link}).

\begin{table}[ht]
\footnotesize
\centering
\begin{tabular}{|c|c|c|c|c|c|c|}
\hline
\textbf{Institution} & \textbf{PheCode} & \textbf{CCS} & \textbf{RxNorm} & \textbf{LOINC} & \textbf{Local Codes} & \textbf{Total} \\ \hline
MGB                 & 1772             & 243          & 1235           & 6370           & 0                   & 9620           \\ \hline
VA                  & 1776             & 224          & 1257           & 1034           & 2673                & 6964           \\ \hline
BCH                 & 1543             & 209          & 1509           & 1942           & 0                   & 5203           \\ \hline
UPMC                & 1841             & 245          & 1987           & 5833           & 8080                & 17986          \\ \hline
MIMIC-IV            & 637              & 129          & 959            & 0              & 2894                & 4619           \\ \hline
\textbf{Total}      & \textbf{1869}    & \textbf{248} & \textbf{4103}  & \textbf{11198} & \textbf{13366}      & \textbf{30784} \\ \hline
\end{tabular}
\caption{Number of unique codes used in five different healthcare systems (MGB, VA, BCH, UPMC, and MIMIC-IV) across the five coding domains: PheCode, CCS, RxNorm, LOINC, and institution-specific local codes.
}
\label{tab:code_summary}
\end{table}

\subsubsection{Evaluating Embedding Quality}
We evaluated the quality of four types of embeddings: (i) PLM-based embeddings, (ii) institution-specific EHR embeddings from the PMI-SVD algorithm, (iii) joint embedding trained with BONMI, and (iv) joint embedding integrated with CODER embeddings (BONMI+) using $w = 0.5$. All embeddings had dimensionality $d = 500$. The quality of embeddings derived from various methods was evaluated in detecting similar or related pairs against random pairs, as described in Step 2.4 of Module 2. Similar or related pairs of codes were selected across four coding domains: PheCode, RxNorm, CCS, and LOINC. The curation of these pairs is described in Section \ref{sec:sim-rel}. 

We also assessed embedding quality in mapping local laboratory codes in the VA data to LOINC codes using $11,808$ curated mappings from OMOP. We reported the top-$k$ accuracy of the codes for each set of embeddings, defined as the proportion of test cases in which the correct mapping for a given code appears among the top-$k$ predictions generated by the embeddings.

\subsection{Downstream Task: Integrative Predictive Modeling}
To highlight the practical utility of PEHRT, we used the trained embeddings from Module 2 to improve the identification and selection of relevant features for predictive modeling. We first focused on predicting eleven diseases: Type 1 Diabetes (T1D), Type 2 Diabetes (T2D), Alzheimer's Disease (AD), Depression (DP), Coronary Atherosclerosis (CA), Congestive Heart Failure (CHF), Congestive Heart Failure - Nonhypertensive (CHFN), Regional Enteritis (RE), Ulcerative Colitis (UC), Rheumatoid Arthritis (RA), and Rheumatoid Arthritis and Other Inflammatory Polyarthropathies (RAO). For each disease, we identified the top $100$ features with the highest cosine similarities to the disease's PheCodes using each embedding method. Additionally, we randomly selected negative features from the complement of the union of features identified by all methods. To evaluate the accuracy of identifying relevant features, we assigned relevance scores (ranging from $0$ to $1$) to each feature using GPT-4. We then computed the AUC for each method, treating the top $100$ features as positive cases and the randomly selected features as negative cases, with the GPT-4 relevance scores serving as probabilities. A higher AUC indicates greater accuracy in selecting relevant features. 

We also considered two additional predictive modeling tasks: predicting future disability status in multiple sclerosis (MS) patients and predicting time to nursing home admission or death in Alzheimer's disease (AD) patients. Both tasks were evaluated at UPMC and MGB using models incorporating demographics (age at baseline, sex, race/ethnicity), healthcare utilization, and selected features using the procedure described in the previous paragraph. For model training, counts of the selected features and the total number of visits, a measure of healthcare utilization, were aggregated over the pre-specified time period at baseline (i.e.,  1 year for predicting future disability status and 2 years for predicting time to nursing home admission or death). We also log-transformed ($x \mapsto \log(x+1)$) the feature counts to improve the stability of model fitting. A lasso-penalized logistic regression model was trained for the disability status outcome and a lasso-penalized Cox proportional hazards model was trained for the time to nursing home admission outcome. The hyperparameter was tuned through five-fold cross-validation. 

\subsection{Results}
\subsubsection{Embedding Validation}
The embedding validation results for detecting known relationship pairs are summarized in Table~\ref{tab:auc_scores}. Overall, PEHRT-based embeddings outperform most PLM-based embeddings in discriminating both similarity and relatedness. Among PLM-based methods, OpenAI and CODER achieve the strongest results, yet they still fall short of BONMI+. This performance gap stems from the fact that PLM-based embeddings are trained primarily on biomedical text corpora and therefore fail to fully capture the nuanced disease patterns and clinical associations present in real-world EHR data. By contrast, the PEHRT-based BONMI+ embedding attains the highest performance on both tasks, benefiting from the representational strengths of PLMs while simultaneously integrating cross-institutional EHR information.

\begin{table}[ht]
\small
\centering
\begin{tabular}{lcccccccc}
\hline
       & BONMI & BONMI+ & CODER & SBERT & BBERT & PBERT & OpenAI & BGE \\
\hline
Similarity  & 0.916 & \textbf{0.966}  & 0.950 & 0.755   & 0.537   & 0.565      & 0.951  & 0.801 \\
Relatedness  & 0.815 & \textbf{0.842}  & 0.811 & 0.682   & 0.477   & 0.547      & 0.832  & 0.690 \\
\hline
\end{tabular}
\caption{Area under the curve (AUC) for similarity and relatedness tasks, reported by method. BONMI and BONMI+ are PEHRT-based embeddings, where BONMI+ further incorporates CODER embeddings. CODER, SBERT (SapBERT), BBERT (BioBERT), PBERT (PubMedBERT), OpenAI, and BGE are PLM-based embeddings pre-trained on biomedical corpora.}
\label{tab:auc_scores}
\end{table}

\begin{table}[!h]
\centering
\small
\begin{tabular}[t]{lrrrrrrrrrr}
\toprule
\multicolumn{1}{c}{ } & \multicolumn{5}{c}{Similarity} & \multicolumn{5}{c}{Relatedness} \\
\cmidrule(l{3pt}r{3pt}){2-6} \cmidrule(l{3pt}r{3pt}){7-11}
Method & MGB & VA & UPMC & BCH & MIMIC & MGB & VA & UPMC & BCH & MIMIC\\
\midrule
PMI-SVD & 0.943 & 0.939 & 0.882 & 0.847 & 0.710 & 0.864 & 0.852 & 0.798 & 0.782 & 0.721\\
BONMI & 0.943 & 0.946 & 0.926 & 0.934 & 0.922 & 0.867 & 0.882 & 0.862 & 0.860 & 0.842\\
BONMI+ & \textbf{0.971} & \textbf{0.969} & \textbf{0.964} & \textbf{0.966} & \textbf{0.961} & \textbf{0.902} & \textbf{0.910} & \textbf{0.902} & \textbf{0.905} & \textbf{0.877}\\
CODER & 0.954 & 0.951 & 0.948 & 0.947 & 0.952 & 0.866 & 0.871 & 0.866 & 0.868 & 0.840\\
SBERT & 0.769 & 0.767 & 0.762 & 0.753 & 0.754 & 0.748 & 0.742 & 0.759 & 0.752 & 0.743\\
BBERT & 0.549 & 0.544 & 0.538 & 0.539 & 0.554 & 0.509 & 0.511 & 0.509 & 0.505 & 0.494\\
PBERT & 0.566 & 0.564 & 0.563 & 0.554 & 0.596 & 0.577 & 0.573 & 0.590 & 0.573 & 0.608\\
OpenAI & 0.952 & 0.948 & 0.950 & 0.944 & 0.952 & 0.861 & 0.865 & 0.864 & 0.865 & 0.817\\
BGE & 0.794 & 0.789 & 0.793 & 0.790 & 0.781 & 0.702 & 0.699 & 0.707 & 0.708 & 0.693\\
\bottomrule
\end{tabular}
\caption{Area under the curve (AUC) for similarity and relatedness tasks, reported by method and institution. PMI-SVD refers to embeddings trained on each institution’s EHR data. BONMI and BONMI+ are PEHRT-based embeddings, where BONMI+ further incorporates CODER embeddings. CODER, SBERT (SapBERT), BBERT (BioBERT), PBERT (PubMedBERT), OpenAI, and BGE are PLM-based embeddings pre-trained on biomedical corpora.}
\label{auc:institu}
\end{table}

Table~\ref{auc:institu} reports institution-specific results, with AUCs calculated using code pairs drawn from each institution separately. As expected, PLM-based embeddings achieve relatively consistent performance across institutions, since they are not trained on institution-specific datasets. In contrast, the PMI-SVD embeddings vary substantially by site, performing reasonably well for large datasets such as MGB, but deteriorating in smaller datasets such as MIMIC-IV. Notably, BONMI and BONMI+ substantially improve performance across all institutions, with BONMI+ consistently achieving the best scores. These gains underscore the advantages of jointly leveraging multi-institutional EHR signals together with semantic information from PLMs.

\newpage

For code mapping, the results in Table~\ref{R1} show that the PLM-based embeddings from SapBERT and OpenAI are superior to BONMI+. This result is expected since code mapping relies heavily on the semantic meaning of code descriptions and it underscores our recommendation to validate the PLM-based embeddings individually as they may be more appropriate for semantically-driven tasks.

\begin{table}[ht]
\centering
\begin{tabular}{lcccccccc}
\hline
       & BONMI & BONMI+ & CODER & SBERT & BBERT & PBERT & OpenAI & BGE \\
\hline
Top-1  & 0.20  & 23.55 & 30.83 & 44.26 & 6.45  & 7.18  & \textbf{49.34} & 34.98 \\
Top-5  & 2.64  & 50.66 & 58.38 & 66.05 & 8.94  & 12.60 & \textbf{79.92} & 52.96 \\
Top-10 & 4.49  & 62.53 & 69.27 & 72.45 & 11.48 & 16.32 & \textbf{85.00} & 59.36 \\
Top-20 & 8.70  & 74.55 & 76.70 & 76.84 & 14.70 & 21.25 & \textbf{88.72} & 65.36 \\
\hline
\end{tabular}
\caption{Top $k$ accuracy of mapping VA local lab codes to LOINC codes using different methods. BONMI and BONMI+ are PEHRT-based embeddings, where BONMI+ further incorporates CODER embeddings. CODER, SBERT (SapBERT), BBERT (BioBERT), PBERT (PubMedBERT), OpenAI, and BGE are PLM-based embeddings pre-trained on biomedical corpora.}
\label{R1}
\end{table}

\subsubsection{Integrative Predictive Modeling}
For the predictive modeling tasks, Table \ref{tab:feature} presents the rank correlations between the cosine similarities of the candidate features and the GPT-4 scores for the 11 target diseases. BONMI+ generally outperforms the other embeddings in selecting features for all diseases. In particular, both BONMI and BONMI+ typically outperform the PLM-based embeddings, as feature selection inherently depends on clinical relationships among EHR codes, which are well documented in real-world EHR data.

\begin{table}[htbp]
\footnotesize 
\centering
\begin{tabular}{lcccccccc}
\toprule
Disease & BONMI & BONMI+ & CODER & SBERT & BBERT & PBERT & OpenAI & BGE \\
\midrule
T1D   & 0.385 & \textbf{0.429} & 0.295 & 0.144 & -0.072 & 0.045 & 0.369 & 0.138 \\
T2D   & 0.479 & \textbf{0.497} & 0.303 & 0.087 & -0.045 & 0.057 & 0.477 & 0.179\\
AD    & 0.313 & 0.362 & 0.289 & 0.164 & -0.079 & 0.021 & \textbf{0.382} & 0.289  \\
DP    & 0.449 & \textbf{0.489} & 0.361 & 0.024 &  0.014 & 0.002 & 0.440 & 0.216  \\
CA    & 0.448 & \textbf{0.478} & 0.343 & 0.055 & -0.028 & 0.033 & 0.426 & 0.007 \\
CHF   & 0.484 & \textbf{0.540} & 0.444 & 0.377 &  0.035 & -0.032 & 0.444 & 0.113  \\
CHFN  & 0.687 & \textbf{0.735} & 0.607 & 0.464 &  0.035 & 0.174 & 0.642 & 0.078 \\
RE    & \textbf{0.289} & 0.252 & 0.115 & 0.059 &  0.080 & 0.005 & 0.206 & 0.107 \\
UC    & 0.262 & 0.215 & 0.067 & 0.048 & -0.006 & 0.029 & \textbf{0.267} & 0.163  \\
RA    & 0.328 & 0.291 & 0.184 & 0.030 &  0.034 & -0.002 & \textbf{0.338} & 0.073  \\
RAO   & \textbf{0.499} & 0.463 & 0.249 & 0.223 &  0.054 & 0.000 & 0.490 & 0.044  \\
AVG. & 0.420 & \textbf{0.432} & 0.296 & 0.152 &  0.002 & 0.030 & 0.407 & 0.128  \\
\bottomrule
\end{tabular}
\caption{Rank correlations between the cosine similarities of the candidate features and the GPT-4 scores for 11 target diseases as well as the average across these diseases. BONMI and BONMI+ are PEHRT-based embeddings, where BONMI+ further incorporates CODER embeddings. CODER, SBERT (SapBERT), BBERT (BioBERT), PBERT (PubMedBERT), OpenAI, and BGE are PLM-based embeddings pre-trained on biomedical corpora.}
\label{tab:feature}
\end{table}

Figures \ref{fig:MS_AUC} and \ref{fig:AD_Cind} present the AUC for the models for MS disability prediction and time to nursing home admission or death for AD patients at MGB and UPMC, respectively. Consistent with our results measuring the quality of feature selection, models incorporating the BONMI and BONMI+ selected features have the strongest performance. Interestingly, models with features selected by institution-specific EHR embeddings achieved better performance than PLM-based embeddings at MGB, but not at UPMC. This finding underscores our recommendation to validate multiple embeddings as results can vary across tasks and institutions.

\begin{figure}[H]
    \centering
    \includegraphics[width=0.8\linewidth]{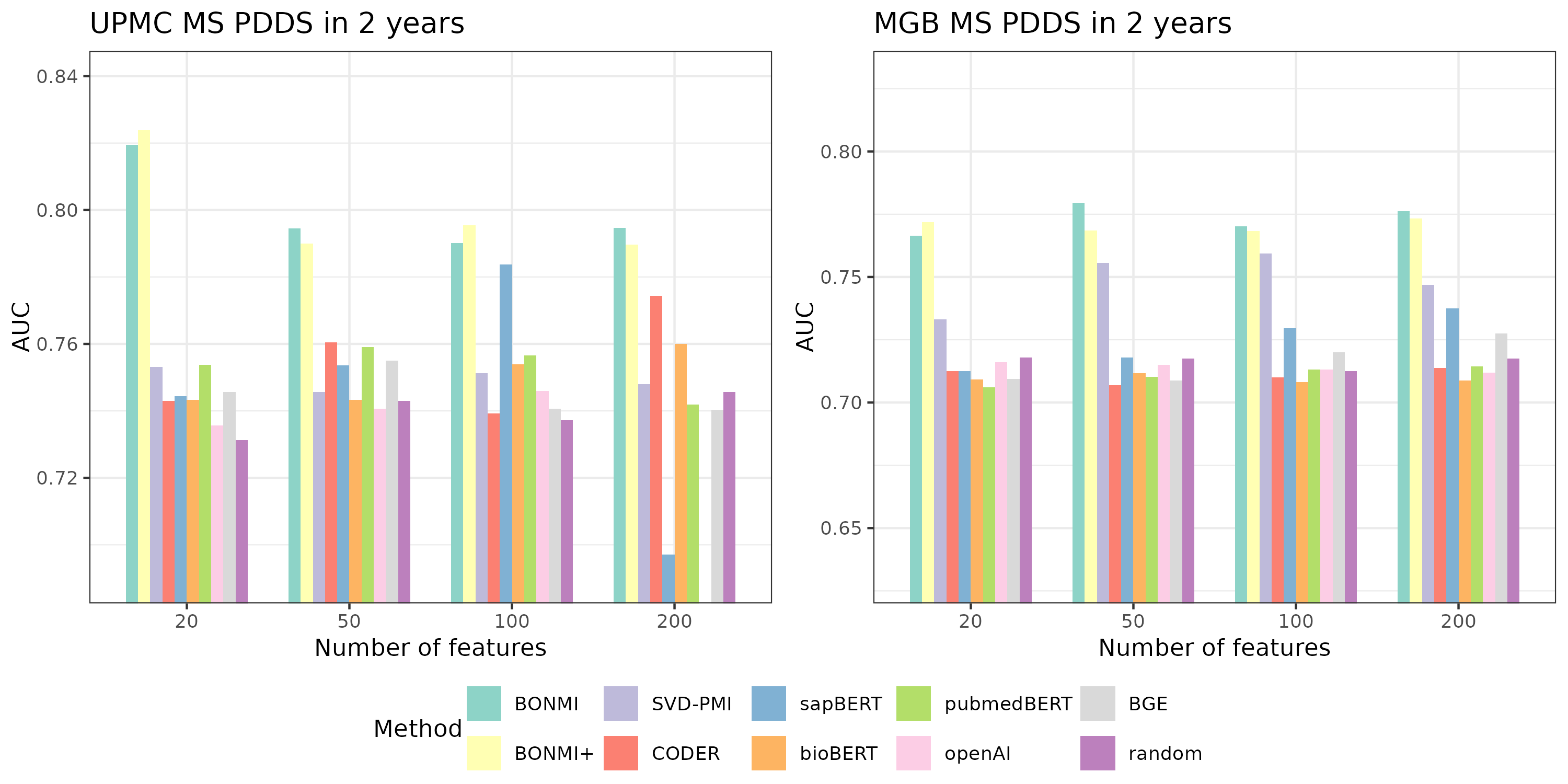}
    \caption{AUC of lasso-penalized logistic regression models for predicting disability status in MS patients based on Patient Determined Disease Steps (PDDS) scores two years after their first visit using varying numbers of selected features (20, 50, 100, and 200). Comparisons are shown for different embedding methods, including BONMI+, BONMI, PLM-based embeddings, institution-specific EHR embeddings (PMI-SVD), and a ``random'' method consisting of randomly selected features combined with the main PheCode and healthcare utilization feature. Results are presented separately for UPMC (left) and MGB (right), with higher AUC values indicate better predictive performance. The training and test sample sizes are both 500.}
    \label{fig:MS_AUC}
\end{figure}

\begin{figure}[H]
    \centering
    \includegraphics[width=0.8\linewidth]{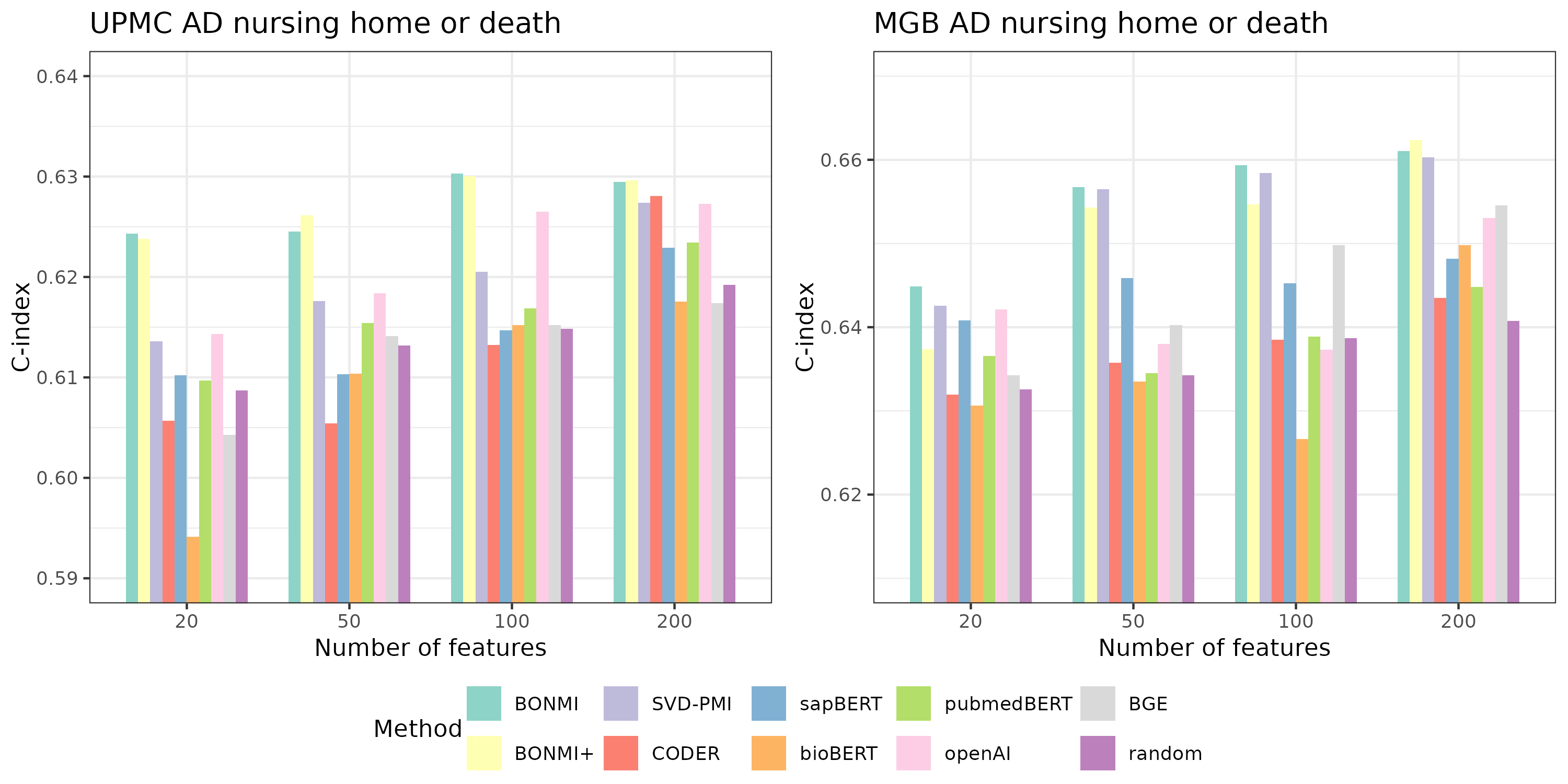}
    \caption{C-index of lasso-penalized Cox proportional hazards models for predicting time to nursing home admission or death in Alzheimer's disease (AD) patients using varying numbers of selected features (20, 50, 100, and 200). Comparisons are shown for different embedding methods, including BONMI+, BONMI, PLM-based embeddings, institution-specific EHR embeddings (PMI-SVD), and a ``random'' method consisting of randomly selected features combined with the main PheCode and healthcare utilization feature. Results are presented separately for UPMC (left) and MGB (right), with higher C-index values indicating better predictive performance. The training sample size is 15,000.}
    \label{fig:AD_Cind}
\end{figure}

\section{Curation of Related \& Similar Pairs}\label{sec:sim-rel}
In UMLS, relationships between biomedical concepts are systematically categorized to facilitate effective information retrieval and integration. Among these, the ``relate'' relationship serves as a broad category encompassing various associative connections between concepts, such as ``cause\_of'', ``may\_treat'', and ``ddx'' (differential diagnosis) \footnote{A detailed description about these relation types can be found in \url{https://www.nlm.nih.gov/research/umls/knowledge_sources/metathesaurus/release/abbreviations.html}.}. The ``similar'' relationship is a specific sub-type that denotes concepts sharing common characteristics or attributes, indicating a high degree of resemblance though not complete identity. For instance, the concepts ``Hypertension'' and ``High Blood Pressure'' are considered similar, as they refer to the same medical condition. In contrast, a general ``relate'' relationship might connect ``Diabetes Mellitus'' and ``Peripheral Neuropathy,'' where the association is based on a causal link (i.e., peripheral neuropathy can be a complication arising from diabetes, rather than direct similarity). 

However, since the UMLS concepts are mostly encoded as CUIs, their relationships cannot be directly translated to relationships for EHR codes. We mapped CUIs to EHR codes using both existing mappings (RxNorm, CCS, LOINC to CUIs) \footnote{\url{https://bioportal.bioontology.org/ontologies}} and GPT-4 (PheCode to CUIs). Specifically, for PheCode to CUIs, we used SapBERT to choose the most similar PheCode for each CUI whose semantic type is ``Disease or Syndrome.'' We then prompted GPT-4 to annotate whether the CUI could be mapped to the selected PheCode and chose the positive pairs as the CUI-PheCode mapping.

We also derived additional pairs from the hierarchical structures of PheCode, RxNorm, and LOINC to capture relationships between similar codes. For PheCodes, more digits indicate greater specificity (e.g., PheCode: 296 for mood disorders is the parent of PheCode: 296.1 for bipolar disorder and PheCode: 296.2 for depression, with PheCode: 296.22 for major depressive disorder as a further refinement). We treated PheCodes sharing the same first three integers as similar relationships. For LOINC, we identified similar relationships that share the same LOINC part parent. For RxNorm leaf codes, we identified similar relationships that share common parents.

\bibliographystyle{apalike}
\bibliography{sample}

\end{document}